\documentclass[letterpaper, 10 pt, conference]{ieeeconf}

\IEEEoverridecommandlockouts                              % This command is only needed if 
\usepackage{graphicx} % for pdf, bitmapped graphics files
\usepackage{epsfig} % for postscript graphics files
\usepackage{dblfloatfix} % fix for bottom-placement of figure
\usepackage{courier}
\usepackage{amsmath} % cmex10
\usepackage{amssymb}
\usepackage{float}
\usepackage{xcolor}
\usepackage{multirow}
\usepackage{caption}
\usepackage{subcaption}
\usepackage{booktabs}
\usepackage[hidelinks]{hyperref}
\usepackage[backend=bibtex,bibstyle=ieee,citestyle=numeric-comp]{biblatex}
\title{\LARGE \bf
Calibration and Uncertainty Characterization for Ultra-Wideband Two-Way-Ranging Measurements
}

\author{Mohammed Ayman Shalaby, Charles Champagne Cossette, James Richard Forbes, Jerome Le Ny\vspace{-15pt}\\  % <-this % stops a space
\thanks{This work was supported by the NSERC Alliance Grant program, by the CFI JELF program, and by the FRQNT.}% <-this % stops a space
\thanks{M. A. Shalaby, C. C. Cossette, and J. R. Forbes are with the department of Mechanical Engineering, McGill University, Montreal, QC H3A 0C3, Canada. {\{\tt\small mohammed.shalaby@mail.mcgill.ca, charles.cossette@mail.mcgill.ca, james.richard.forbes@mcgill.ca\}.} J. Le Ny is with the department of Electrical Engineering, Polytechnique Montreal, Montreal, QC H3T 1J4, Canada. \{\tt\small jerome.le-ny@polymtl.ca\}.}}

\newcommand{\ignore}[1]{}

\newcommand{\norm}[1]{\left\Vert#1\right\Vert} % Norm
\newcommand{\bma}[1]{\left[\begin{array}{ #1}}
\newcommand{\ema}{\end{array}\right]}

\DeclareMathAlphabet{\mbf}{OT1}{ptm}{b}{n}

\newcommand{\mbfhat}[1]{{\hat{\mbf{#1}}}}

\def\fdotb{{\raisebox{-0.6ex}{ \kern0.2ex\raisebox{0.8ex}{\tiny $\hspace*{-1ex}\circ$}}}}
\def\fddotb{{\raisebox{-0.6ex}{ \kern0.2ex\raisebox{0.8ex}{\tiny $\hspace*{-1ex}\circ\circ$}}}}

\newcommand{\f}{\frac}

\newcommand{\trans}{{\ensuremath{\mathsf{T}}}} % transpose
\newcommand{\utimes}{ {\raisebox{-0.6ex}{ \kern-1.0ex\raisebox{0.6ex}{ \small $\mathsf{v}$}}} } % 
\newcommand{\beq}{\begin{equation}}
\newcommand{\eeq}{\end{equation}}
\newcommand{\bdis}{\begin{displaymath}}
\newcommand{\edis}{\end{displaymath}}
\newcommand{\beqarray}{\begin{eqnarray}}
\newcommand{\eeqarray}{\end{eqnarray}}
\newcommand{\beqarraynn}{\begin{eqnarray*}}
\newcommand{\eeqarraynn}{\end{eqnarray*}}
\newcommand{\balign}{\begin{align}}
\newcommand{\ealign}{\end{align}}
\newcommand{\balignnn}{\begin{align*}}
\newcommand{\ealignnn}{\end{align}}

\renewcommand{\theenumii}{\arabic{enumii}}
\makeatletter
\renewcommand{\p@enumii}{\theenumi.}
\makeatother
\usepackage{arydshln}

\begin{document}

\maketitle
\thispagestyle{empty}
\pagestyle{empty}

%%%%%%%%%%%%%%%%%%%%%%%%%%%%%%%%%%%%%%%%%%%%%%%%%%%%%%%%%%%%%%%%%%%%%%%%%%%%%%%%
\begin{abstract}
   Ultra-Wideband (UWB) systems are becoming increasingly popular for indoor localization, where range measurements are obtained by measuring the time-of-flight of radio signals. However, the range measurements typically suffer from a systematic error or bias that must be corrected for high-accuracy localization. In this paper, a ranging protocol is proposed alongside a robust and scalable antenna-delay calibration procedure to accurately and efficiently calibrate antenna delays for many UWB tags. Additionally, the bias and uncertainty of the measurements are modelled as a function of the received-signal power. The full calibration procedure is presented using experimental training data of 3 aerial robots fitted with 2 UWB tags each, and then evaluated on 2 test experiments. A localization problem is then formulated on the experimental test data, and the calibrated measurements and their modelled uncertainty are fed into an extended Kalman filter (EKF). The proposed calibration is shown to yield an average of 46\% improvement in localization accuracy. Lastly, the paper is accompanied by an open-source UWB-calibration Python library, which can be found at \href{https://github.com/decargroup/uwb_calibration}{https://github.com/decargroup/uwb\_calibration}.
\end{abstract}
%%%%%%%%%%%%%%%%%%%%%%%%%%%%%%%%%%%%%%%%%%%%%%%%%%%%%%%%%%%%%%%%%%%%%%%%%%%%%%%%
\section{Introduction}

Robotic localization and mapping applications typically require a means of acquiring position information relative to a reference point with known location. Global Navigation Satellite System (GNSS) provides accurate and precise positioning information outdoors; however, localization performance degrades significantly in obstructed or indoor environments \cite{Groves2011, Irish2014a}. An attractive indoor localization option that has been increasingly gaining traction is the use of \emph{ultra-wideband} (UWB) radio signals between transceivers, or \emph{tags}, as a means of ranging. UWB transceivers, such as the DWM1000 module provided by Decawave \cite{dw1000}, are typically inexpensive, consume little power, and provide a means for data transfer between robots, thus deeming them particularly useful for a variety of robotic applications \cite{Mueller2015a, Hepp2016a, Cao2020}.

\begin{figure}[t!]
   \centering
   \begin{subfigure}[t]{0.43\columnwidth}
       \centering
       \includegraphics[width=\columnwidth]{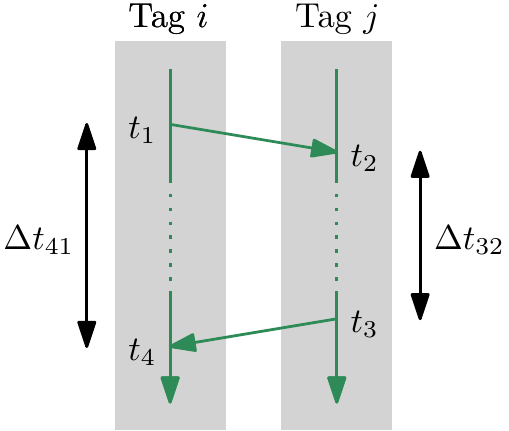}
       \caption{SS-TWR.}
       \label{fig:ss_twr}
   \end{subfigure}%
   ~ \hspace{2pt}
   \begin{subfigure}[t]{0.43\columnwidth}
       \centering
       \includegraphics[width=\columnwidth]{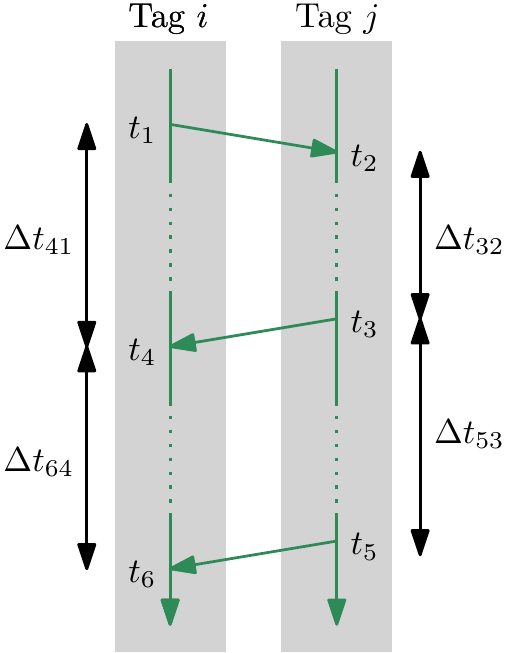}
       \caption{Proposed DS-TWR.}
       \label{fig:ds_twr}
   \end{subfigure}
   \caption{Timeline schematics for two tags $i$ and $j$ representing the different TWR ranging protocols, where $t_\ell$ represents the $\ell^\mathrm{th}$ timestamp for a TWR instance and $\Delta t_{\ell k} \triangleq t_\ell - t_k$.}
   \vspace{-12pt}
\end{figure}

UWB-based ranging typically relies on measuring the \emph{time-of-flight} (ToF) of radio signals from one tag to another. This requires estimating the offset between the clock on each tag. Furthermore, the clocks often run at different rates due to physical imperfections in the individual clock's crystal oscillator, causing the offset to be time-varying. The rate of change of the clock offset is referred to as the clock \emph{skew}. In order to negate the effect of the clock offset during ranging, different ranging protocols have been proposed, with the choice being dependent on the specific application and availability of tags \cite{sahinoglu2008}, \cite[Section 7.1.4]{groves2013}. A commonly used protocol is \emph{two-way ranging} (TWR), which relies on averaging out the measured ToF between two signals to negate the clock offset. This form of TWR is referred to as \emph{single-sided} TWR (SS-TWR), and is shown in Figure \ref{fig:ss_twr}.

Nonetheless, even after correcting for clock offsets, UWB range measurements typically suffer from a systematic error or \emph{bias}. A significant contributor to this error is the skew between the clocks of the two ranging tags, as the different tags measure the passage of time in different units \cite{Cano2022, Neirynck2017}. This additional bias can be corrected by estimating the clock skew between the tags and embedding a skew-dependent correction factor when computing the range measurement, as proposed in \cite{Cano2022}. However, this necessitates estimating the clock skew between all tags involved in ranging. Alternatively, \cite{Neirynck2017} proposes a form of computing the range measurement utilizing \emph{double-sided} TWR (DS-TWR), which is shown to mitigate clock-skew-dependent bias.

Another source of ranging bias stems from relative-pose-dependent antenna radiation pattern \cite{DecawaveSourcesOfError}, where pose refers to both position and attitude. The varying signal strength can cause timestamping errors, and this effect is typically addressed using data-driven models. In \cite{Gonzalez2009}, a simple experiment with pre-localized fixed tags or \emph{anchors} is used to determine a relation between bias and the distance between ranging tags, while in \cite{Wymeersch2012}, models are trained using the distance between the tags and 7 features extracted from the channel impulse response (CIR). In \cite{Ledergerber2018} and \cite{Zhu2019}, a robot is flown around in a room with UWB anchors to learn a model of the range bias as a function of the robot's pose. The main drawback of these methods is that the learned model is dependent on the relative poses of the ranging tags, which are typically unknown in real-time without the bias-corrected measurements in the first place. Additionally, the learnt models are trained and tested on the same anchor formations and are therefore not necessarily generalizable; calibration must occur for every new anchor formation. In \cite{Cano2022}, the former issue is addressed by finding a relation between the bias and the received \emph{first-path power} (FPP) in \emph{line-of-sight} (LOS) conditions with 2D motion.

Delays in communication between the embedded microchip and the UWB antenna are another source of bias \cite{DevawaveAntennaDelay}. This antenna delay is roughly the same for different UWB tags with the same physical design and is at least a few hundreds of nanoseconds \cite{DevawaveAntennaDelay}, but can vary tenths of a nanosecond or more from tag to tag due to manufacturing inaccuracies. Given that a one-nanosecond timestamping error corresponds to 30 cm in ranging error, the need to perform antenna-delay calibration for every tag is critical. In \cite{DevawaveAntennaDelay}, a basic TWR-based calibration procedure is suggested for calibrating antenna delays. However, the lack of motion introduces a risk of learning the aforementioned relative-pose-dependent bias as antenna delays. In \cite{Cano2022}, experiments involving a pair of tags at a time ranging with each other is used to fit what is referred to as a ``pair-dependent constant''. Therefore, the calibration procedure involves calibrating the relative delay between one pair at a time, which does not scale well to systems with many UWB tags.

This paper addresses the problem of calibrating UWB tags, and the main contributions are as follows. 
\begin{itemize}
   \item An alternative DS-TWR protocol is proposed and is shown to mitigate the clock-skew-induced bias.
   \item A scalable antenna-delay calibration algorithm is presented that is robust to outliers and pose-dependent bias.
   \item The bias-versus-FPP fit presented in \cite{Cano2022} is extended to also address the uncertainty of the measurements as a function of FPP, and DS-TWR is utilized to overcome the need to estimate the clock skew.
   \item The proposed antenna-delay and bias-FPP calibration are evaluated on an aerial experiment with no anchors, where all the tags are fitted to moving robots. 
   \item The code for the full calibration procedure is attached to this paper as an open-access online repository, which can be found at \href{https://github.com/decargroup/uwb_calibration}{https://github.com/decargroup/uwb\_calibration}.
\end{itemize}

The remainder of this paper is organized as follows. The proposed DS-TWR is discussed in Section \ref{sec:ranging_protocol}, alongside a theoretical analysis of the clock-skew-dependent bias. In Section \ref{sec:antenna_delay_calibration}, a robust antenna-delay calibration algorithm is presented, followed by the bias and uncertainty calibration as a function of FPP in Section \ref{sec:power_correlated_calibration}. The calibration methods presented in Sections \ref{sec:antenna_delay_calibration} and \ref{sec:power_correlated_calibration} are introduced on the same experimental training data, and are then evaluated on 2 testing experiments in Section \ref{sec:experimental_results}.

\section{The Ranging Protocol} \label{sec:ranging_protocol}

UWB ranging relies on the time-of-flight (ToF) of signals between two tags in order to compute range measurements. The simplest way to do this is using SS-TWR, shown in Figure~\ref{fig:ss_twr}, where the ToF measurement can be computed as 
\beq
\label{eq:ss_twr}
   t_\mathrm{f} = \f{1}{2} (\Delta t_{41} - \Delta t_{32}).
\eeq
However, different UWB tags have differrent clocks that are typically running at different rates, and this clock skew results in additional bias in the computed ToF measurement. In \cite{Neirynck2017}, an alternative DS-TWR-based ranging protocol is proposed to mitigate clock-skew-dependent bias. In this paper, the DS-TWR protocol shown in Figure~\ref{fig:ds_twr} is proposed, which differs from \cite{Neirynck2017} by having the responding tag instead of the initiating tag transmit the third signal. The ToF measurement can then be computed as 
\beq
   \label{eq:ds_twr}
   t_\mathrm{f} = \f{1}{2} \left(\Delta t_{41} - \f{\Delta t_{64}}{\Delta t_{53}} \Delta t_{32} \right).
\eeq
This protocol is motivated by the intuitive understanding that the additional correcting factor in \eqref{eq:ds_twr} transforms $\Delta t_{32}$ from time units of the receiver tag's clock to time units of the initiator tag's clock. Additionally, the proposed ranging protocol allows the initiating tag to process the range measurement by computing \eqref{eq:ds_twr}, without requiring additional signals for the responding tag to send $\Delta t_{32}$ and $\Delta t_{53}$.

\subsection{Analytical Bias Model} \label{subsec:analyital_bias_model}

To demonstrate clock-skew-dependent bias, consider in SS-TWR the clock-skew-corrupted ToF measurement,
\beq
   \tilde{t}_\mathrm{f}^\mathrm{ss} = \f{1}{2} \left( (1+\gamma_i) (\Delta t_{41} + \eta_{41}) - (1+\gamma_j) (\Delta t_{32} + \eta_{32})  \right), \label{eq:t_f_ss}
\eeq
where $\gamma_i$ is the skew of Tag $i$'s clock relative to real time, $\eta_{k \ell} = \eta_{k} - \eta_{\ell}$, and $\eta_k, \eta_\ell \sim \mathcal{N}(0,R)$ are mutually-independent timestamping white noise associated with timestamps $t_k$ and $t_\ell$, respectively. The ToF error is thus
\begin{align}
   e^\mathrm{ss} &\triangleq \tilde{t}_\mathrm{f}^\mathrm{ss} - t_\mathrm{f} \nonumber \\
   &= \f{1}{2} \left( \gamma_i \Delta t_{41} + (1+\gamma_i) \eta_{41} - \gamma_j \Delta t_{32} - (1+\gamma_j) \eta_{32} \right), \label{eq:e_ss}
\end{align}
and the expected value of $e^\mathrm{ss}$ is  
\begin{align}
   \mathbb{E} \left[ e^\mathrm{ss} \right] &= \f{1}{2} \left( \gamma_i \Delta t_{41} - \gamma_j \Delta t_{32} \right) \nonumber\\
   &\stackrel{\eqref{eq:ss_twr}}{=} \f{1}{2} \left( \gamma_i (2 t_\mathrm{f} + \Delta t_{32}) - \gamma_j \Delta t_{32} \right) \nonumber\\
   &= \gamma_i t_\mathrm{f} + \f{1}{2} \left( \gamma_i  - \gamma_j \right) \Delta t_{32}. \label{eq:ss_mean}
\end{align}

The first component of \eqref{eq:ss_mean} is negligible as skew is in the order of parts-per-million and ToF in nanoseconds. However, $\Delta t_{32}$ is typically in hundreds of microseconds, meaning that clock-skew-dependent bias is not negligible. 

Negating the second component of \eqref{eq:ss_mean} is the motivation behind the proposed ranging protocol. Rewriting \eqref{eq:ds_twr} as
\begin{align}
   t_\mathrm{f} = \f{\Delta t_{41}\Delta t_{53} - \Delta t_{64}\Delta t_{32}}{2 \Delta t_{53}}, \nonumber
\end{align}
and following the same steps as in \eqref{eq:t_f_ss}-\eqref{eq:e_ss}, the ToF error for the proposed DS-TWR can be derived to be approximately
\begin{align}
   e^\mathrm{ds} &\triangleq \tilde{t}_\mathrm{f}^\mathrm{ds} - t_\mathrm{f} \nonumber \\
   &\approx \gamma_i t_\mathrm{f} + \f{1}{2} \left( 1 + \gamma_i \right) \left[ \f{\Delta t_{32}}{\Delta t_{53}} (\eta_{53} - \eta_{64}) + \eta_{41} - \eta_{32} \right],  \label{eq:e_ds}
\end{align}
\vspace{-10pt}

\noindent where $\tilde{t}_\mathrm{f}^\mathrm{ds}$ is the clock-skew-corrupted time-of-flight measurement. Deriving \eqref{eq:e_ds} relies on the assumptions that $\Delta t_{53} \gg \norm{\eta_{53}}$, $\Delta t_{32} \gg \norm{\eta_{32}}$.
This is expected since the timestamping error is typically in the order of nanoseconds or less, and the delay intervals are in the order of hundreds of microseconds. The expected value of the error is therefore
\begin{align}
   \mathbb{E} \left[ e^\mathrm{ds} \right] =  \gamma_i t_\mathrm{f}, \nonumber
\end{align}
which suffers from less bias when compared to the error of the SS-TWR protocol.

\section{Antenna-Delay Calibration} \label{sec:antenna_delay_calibration}

The delay between a chip timestamping transmission and the antenna actually transmitting the signal is referred to as the antenna transmission delay $d^\mathrm{t}$, while the delay between an antenna receiving a signal and the chip timestamping reception is the antenna reception delay $d^\mathrm{r}$. Looking back at Figure~\ref{fig:ds_twr}, the measured time-stamps are therefore
\begin{align}
    \tilde{t}_k &= t_k + d^\mathrm{t}, \qquad k \in \{1,3,5\}, \label{eq:transmission_delay} \\
    \tilde{t}_\ell &= t_\ell + d^\mathrm{r}, \qquad \ell \in \{2,4,6\}. \label{eq:reception_delay}
\end{align}

In this section, a scalable antenna-delay calibration procedure is presented that addresses the need for incorporating motion. In particular, a linear least-squares approach is presented utilizing DS-TWR, which is solved using robust least squares to accommodate for outliers \cite[Section 5.3.2]{barfoot_2017}. Both the antenna delays and the presented approach are environment independent, and therefore the antenna-delay calibration only needs to be performed once for new uncalibrated transceivers.

\subsection{Least Squares Formulation}

The goal of antenna-delay calibration is to find the best-fit delays based on some collected data. In order to perform this calibration procedure, the tags to be calibrated must be capable of DS-TWR. The effect of antenna delays on the DS-TWR ToF measurements is shown by substituting \eqref{eq:transmission_delay} and \eqref{eq:reception_delay} into \eqref{eq:ds_twr}, thus yielding 
\vspace{-3pt}
\begin{align}
    t_\mathrm{f} &= \f{1}{2} \Big( \Delta \tilde{t}_{41} + \underbrace{ d_i^\mathrm{t} - d_i^\mathrm{r} }_{d_i} - \f{ \Delta \tilde{t}_{64}}{\Delta \tilde{t}_{53}} \big( \Delta \tilde{t}_{32}  \underbrace{ - d_j^\mathrm{t} + d_j^\mathrm{r}}_{-d_j} \big) \Big) \label{eq:tof_with_delays}
\end{align}
\vspace{-10pt}

\noindent when Tag $i$ initiates with Tag $j$, where $d_i^\mathrm{t}$ and $d_i^\mathrm{r}$ are the antenna transmission and reception delays of Tag $i$, respectively. In this case, transmission and reception delays can be combined into one delay variable $d_i$ to be estimated for every tag, where $d_i = d_i^\mathrm{t} - d_i^\mathrm{r}$. This is sufficient for systems where only TWR is utilized, which is the focus of this paper. When other ranging protocols are implemented such as time-difference-of-arrival (TDoA) or time-of-arrival (ToA), another antenna-delay calibration procedure is necessary to solve for $d_i^\mathrm{t}$ and $d_i^\mathrm{r}$ separately.

In the presence of $n$ tags to be calibrated, let $\mathcal{P}$ denote the ordered set of tuples representing all ranging pairs of tags. Consequently, the antenna delays are calibrated by formulating a linear least-squares problem as \vspace{-3pt}
\beq
    \mbfhat{d} = \arg \min_{\mbf{d} \in \mathbb{R}^n} \sum_{(i,j) \in \mathcal{P}} \sum_{k=1}^{m_{ij}} g\left(e_{ij}^k(\mbf{d})\right), \label{eq:delay_ls_problem}
\eeq
\vspace{-7pt}

\noindent where the error $e_{ij}^k$ is defined from \eqref{eq:tof_with_delays} as \vspace{-5pt}
\beq
e_{ij}^k(\mbf{d}) = \f{1}{2}(d_i + K^k d_j) - t_f^k + \f{1}{2}(\Delta \tilde{t}_{41}^k - K^k \Delta \tilde{t}_{32}^k), \nonumber
\eeq
\vspace{-12pt}

\noindent $m_{ij}$ is the number of range measurements between Tags $i$ and $j$, $\mbf{d} = \bma{ccc} \hspace{-2pt} d_1 \hspace{-3pt} & \hspace{-3pt} \cdots \hspace{-2pt} & \hspace{-2pt} d_n \hspace{-2pt} \ema^\trans$, the superscript $k$ denotes the $k^\mathrm{th}$ measurement, and $K^k \triangleq \Delta \tilde{t}_{64}^k / \Delta \tilde{t}_{53}^k$. Moreover, $g$ is the loss function, and the choice of $g$ is discussed in Section \ref{subsec:antenna_delay_training_exp}.

If $n=2$, the formulated least-squares problem would have 2 unknowns and only 1 pair of ranging tags, which results in non-uniqueness of the solution. Therefore, the calibration procedure should involve at least 3 tags, yielding 3 unknowns and 3 pairs of ranging tags.

\subsection{Experimental Results on Training Data} \label{subsec:antenna_delay_training_exp}

\begin{figure}[t!]
   \centering
   \begin{subfigure}[t]{0.4\columnwidth}
      \centering
      \includegraphics[width=\columnwidth]{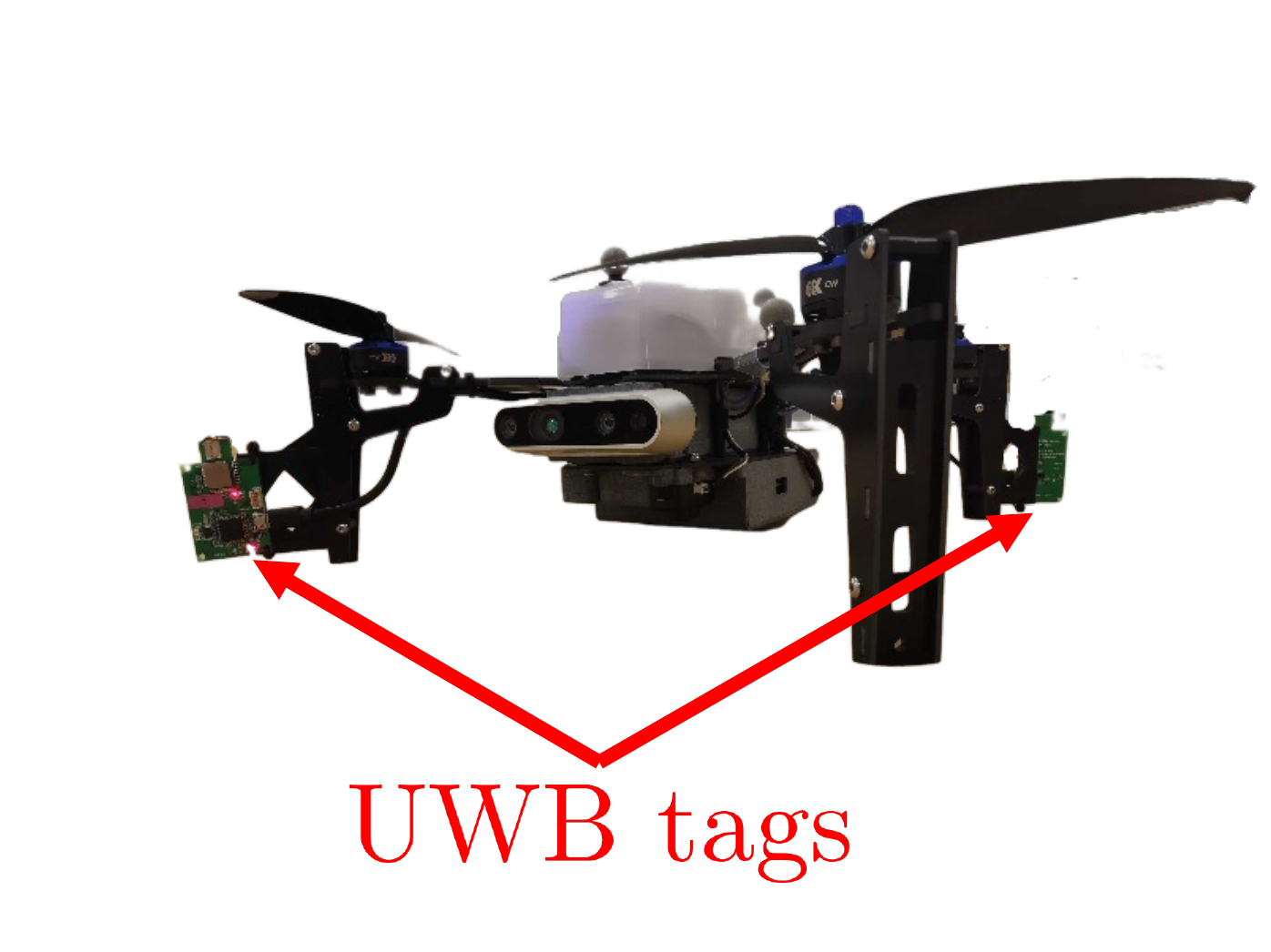}
      \caption{Quadcopter equipped with two UWB tags.}
      \label{fig:drone_with_tags}
   \end{subfigure}%
   ~
   \begin{subfigure}[t]{0.6\columnwidth}
      \centering
      \includegraphics[trim={0cm 0cm 0cm 0cm},clip,width=\columnwidth]{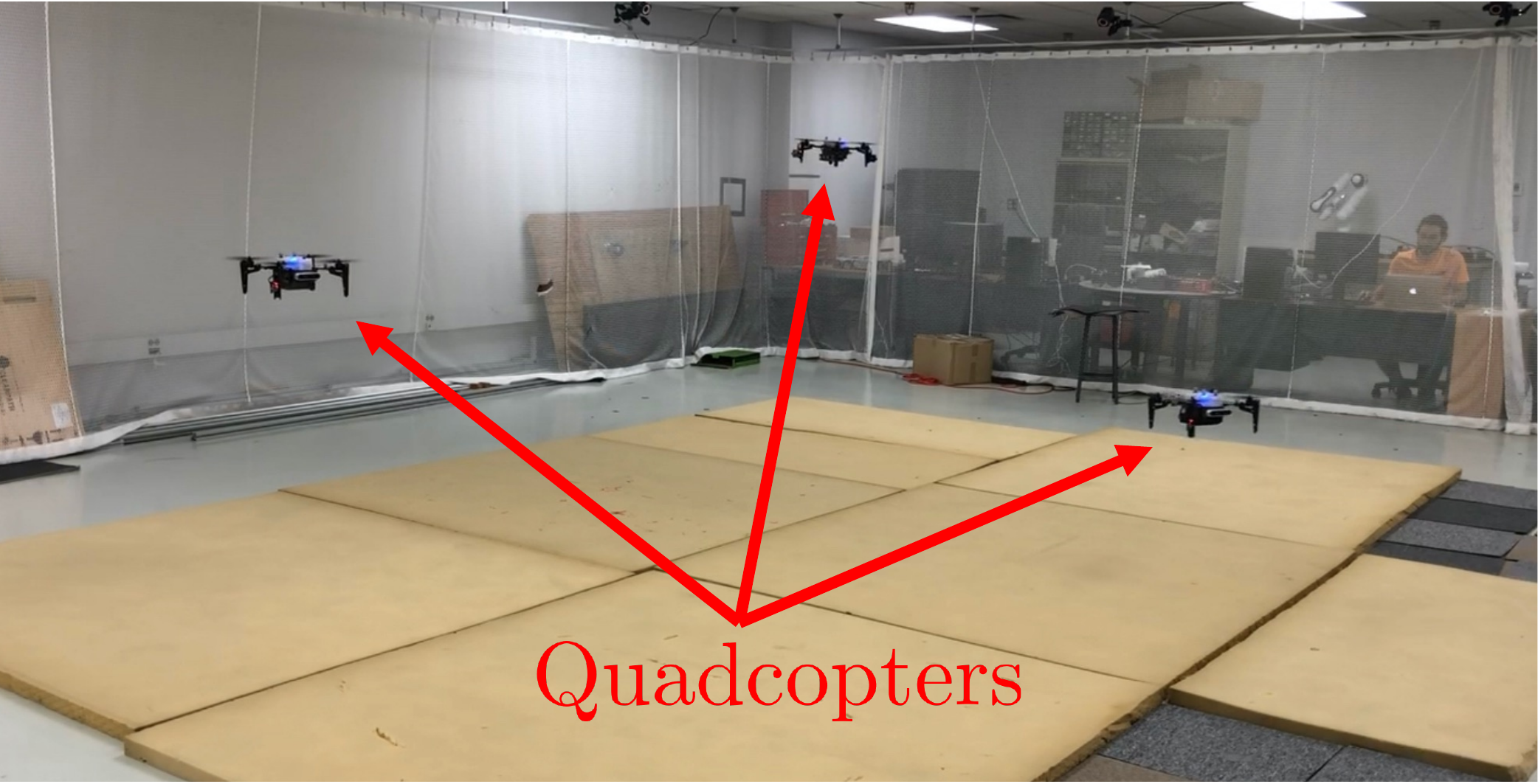}
      \caption{Snapshot from training experiment.}
      \label{fig:flying_drones}
   \end{subfigure}
   \caption{Experimental set-up for antenna-delay and bias-FPP calibration.}
   \vspace{-12pt}
\end{figure}

To evaluate the proposed antenna-delay calibration procedure experimentally, three Uvify IFO-S quadcopters are equipped with 2 UWB tags each as shown in Figure~\ref{fig:drone_with_tags}, for a total of 6 tags. In order to compute the theoretical ToF $t_f$ for any measurement to formulate a similar problem to \eqref{eq:delay_ls_problem}, a motion-capture system is used to get the ground truth distances between the ranging tags. A motion-capture system is chosen for its mm-accuracy, but any other localization approach would suffice, with the accuracy of the calibration depending on the accuracy of the localization algorithm. Unlike the static experiments suggested in \cite{DevawaveAntennaDelay}, this allows a dynamic experiment where the quadcopters fly randomly in 3-dimensional space as shown in Figure~\ref{fig:flying_drones}, which reduces the proneness to learning relative-pose-dependent biases. This dataset consists of 4 minutes of flight time and a total of 38000 range measurements. The calibration procedure can be done by fitting each drone with one tag, but it is common in localization problems to fit two tags to overcome the lack of bearing information \cite{Hepp2016a,Nguyen2018,Shalaby2021}. Nonetheless, the ranging schedule does not directly involve TWR measurements between any two tags on the same drone. Therefore, there are 6 unknown delays and 12 pairs of ranging tags. 

Typically, problems of the form \eqref{eq:delay_ls_problem} are solved by finding $\mbf{d}$ that minimizes the squared error (i.e., choosing $g$ to be L2 loss), which is derived from an assumption that the underlying distribution of the noise is Gaussian. However, UWB measurements suffer from positive outliers due to multipath propagation and other sources of error, and are better modelled using Cauchy distributions \cite{Kok2015}. It is therefore proposed for this particular application to minimize the Cauchy loss $g(x) = \operatorname{log} \left( 0.5 x^2 + 1 \right)$
instead to solve \eqref{eq:delay_ls_problem} while reducing the effect of outliers \cite[Section 5.3.2]{barfoot_2017}.

\begin{figure}
   \includegraphics[width=\columnwidth]{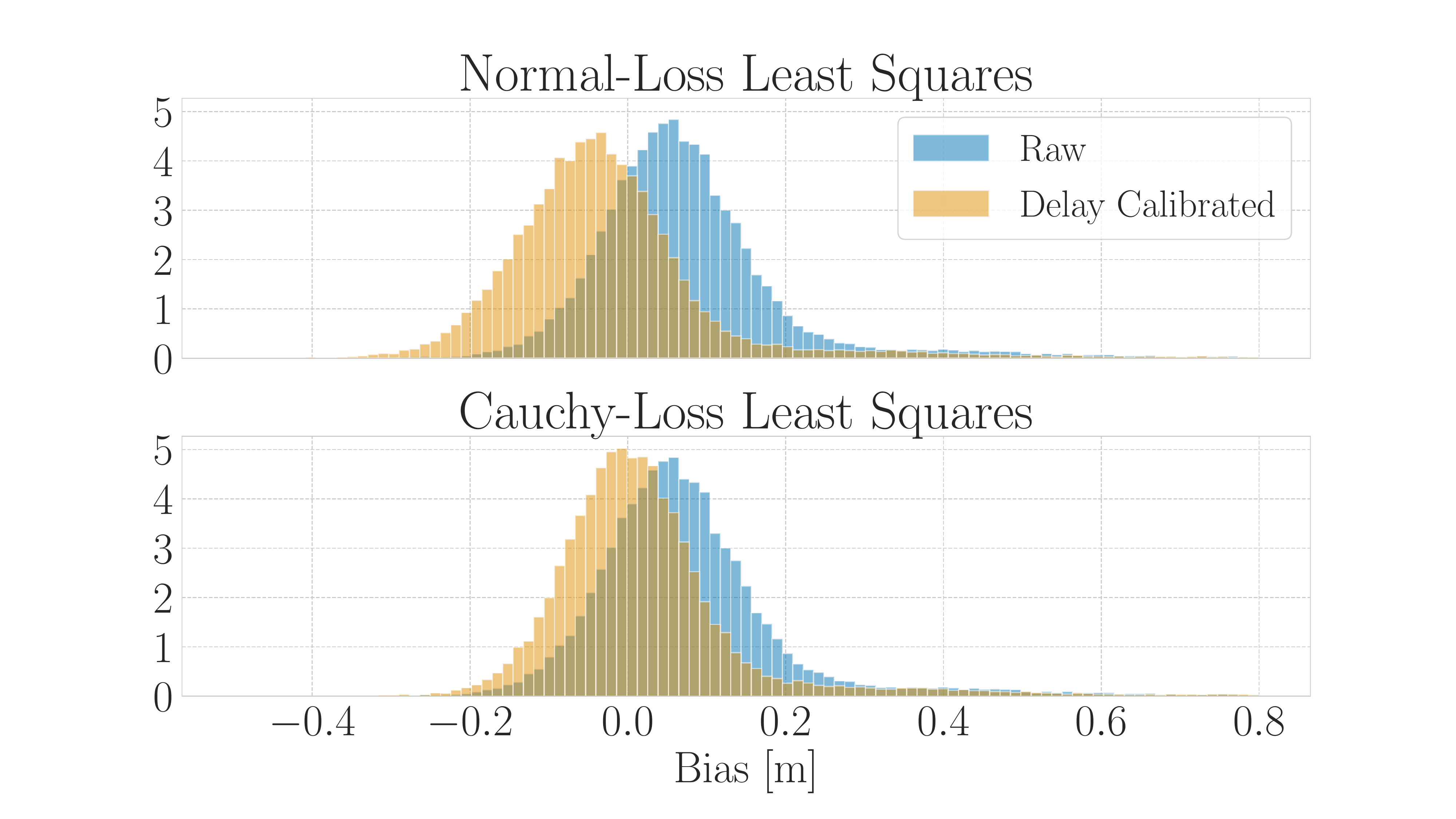}
   \caption{Histograms showing the effect on the ranging bias post-antenna-delay calibration. (Top) Using L2 loss. (Bottom) Using Cauchy loss.}
   \label{fig:normal_vs_cauchy}
   \vspace{-12pt}
\end{figure}

The use of the Cauchy loss is found to be useful for mitigating the effect of positive outliers as the mode of the bias distribution becomes 0, as shown in Figure~\ref{fig:normal_vs_cauchy}. Moreover, the precision of the proposed approach is evaluated by comparing antenna-delay solutions derived from 2 datasets collected one month apart, and the delays converge to the same result within 0.03 ns of accuracy, which corresponds to less than 1 cm of ranging error. %, as shown in Table~\ref{tab:antenna_delay_over_time}. 

% \begin{table}
%    \centering
%    \caption{Calibrated antenna delays for all 6 tags using two training experiments collected 30 days apart. Tag 2 is replaced with Tag 7 in the latter experiment due to incurred physical damage.}
%    \label{tab:antenna_delay_over_time}
%    \begin{tabular}{ c||c|c } 
%     Tag ID & $d$ at day 0 [ns] & $d$ at day 30 [ns] \\ 
%     \hline\hline
%     1 & -0.3638 & -0.3359 \\ 
%     2 & -0.0072 & $\times$ \\ 
%     3 & -0.0688 & -0.0686 \\ 
%     4 & -0.3615 & -0.3799 \\ 
%     5 & -0.1665 & -0.1651 \\ 
%     6 & -0.1163 & -0.0875 \\ 
%     7 & $\times$ & 0.2377 
%    \end{tabular}
% \end{table}

\subsection{Using Calibrated Tags to Calibrate New Tags}

In \cite{Cano2022}, the sum of the antenna delays for every pair of tags $(i,j)$ is calibrated as one constant $c_{ij}$, which for the DS-TWR protocol would be of the form $c_{ij} = d_i + K d_j$.
This approach requires that calibration is done for every pair of tags, which is tedious and not scalable. That is due to the fact that lumping the delay terms into one constant fails to utilize the constant delay terms that appear in different pairs.

By solving for the aggregate antenna delays $d_i$ and $d_j$ individually, this allows calibrating a new tag without collecting data between the new tag and all previously calibrated tags. In order to calibrate a new tag Tag $j$, only one calibrated tag Tag $i$ is required, which then allows solving for $d_j$ from \eqref{eq:tof_with_delays},
\beq
   \hat{d}_j = \f{2t_\mathrm{f} - \Delta \tilde{t}_{41} - \hat{d}_i}{K} + \Delta \tilde{t}_{32}. \nonumber
\eeq
However, it is still recommended to collect more data using a dynamic experiment with the two tags in order to improve robustness to noise and pose-dependent bias. % A comparison between the number of tags required to perform calibration using pair-dependent constants as in \cite{Cano2022} and using the proposed approach is shown in Table~\ref{tab:tags_for_calib}.

% \begin{table}
%    \caption{Comparison of the delays calibrated and the minimum number of tags required for calibration.}
%    \label{tab:tags_for_calib}
%    \begin{tabular}{ c||p{1.5cm}|p{1.5cm}|p{2.6cm} } 
%      & Delays Calibrated & Min. num. of tags & Num. of tags to calibrate 1 tag \\ 
%     \hline\hline
%     \cite{Cano2022} & $d_i + Kd_j$ & 2 & All calibrated tags \\ 
%     Proposed & $d_i$, $d_j$ & 3 & 1 calibrated tag \\ 
%    \end{tabular}
% \end{table}

\section{Power-Correlated Calibration} \label{sec:power_correlated_calibration}

 Another source of error in UWB-based ranging is irregularities in the antenna radiation pattern and system design elements, such as PCB-induced losses. Typically, such losses introduce biases in the measurements that are pose-dependent and that are correlated with the received signal power \cite{DecawaveSourcesOfError,Cano2022}. In this section, the experiments in \cite{Cano2022} are extended in the following ways.
 \begin{enumerate}
    \item The proposed DS-TWR is used rather than SS-TWR, which overcomes the need to estimate the skew between all pairs of tags.
    \item The results are shown to hold for experiments in three-dimensional space.
    \item The results are shown to hold for experiments with some non-LOS measurements due to occlusions from the quadcopters' bodies.
    \item The individual measurements are used in the data-fitting process rather than averaging out measurements from a discrete number of relative poses. 
 \end{enumerate}
 The last point is particularly important as it overcomes the need to remain static during data collection, which allows the calibration procedure to be a simple experiment of robots moving randomly and covering as many relative poses as possible in a relatively short period of time. Another advantage of using all the data in the calibration process is that there is no loss of variance information through averaging out similar measurements. Consequently, the relation between the variance of the measurements and the received signal power can then be analyzed.
 
 \subsection{Bias Calibration}
 
 The bias calibration procedure is similar to the one presented in \cite{Cano2022}. The reception timestamp at Tag $i$ is usually corrupted by an unknown function $\rho_i(\cdot)$ of the received FPP $p^\mathrm{f}$; therefore, from \eqref{eq:ds_twr}, 
 \begin{align}
    t_\mathrm{f} &= \f{1}{2} \Big( \Delta \tilde{t}_{41} + \rho_i\left(p_4^\mathrm{f}\right) \nonumber \\ &\hspace{30pt} - \f{\Delta \tilde{t}_{64}}{\Delta \tilde{t}_{53} + \rho_i\left(p_6^\mathrm{f}\right) - \rho_i\left(p_4^\mathrm{f}\right)} \left( \Delta \tilde{t}_{64} - \rho_j\left(p_2^\mathrm{f}\right) \right) \Big) \nonumber\\
    &\stackrel{(\mathrm{a})}{\approx} \f{1}{2} \left( \Delta \tilde{t}_{41} - \f{\Delta \tilde{t}_{64}}{\Delta \tilde{t}_{53}} \Delta \tilde{t}_{64} \right) + \f{1}{2} \left( \rho_i\left(p_4^\mathrm{f}\right) + \rho_j\left(p_2^\mathrm{f}\right) \right) \nonumber\\
    &\triangleq \f{1}{2} \left( \Delta \tilde{t}_{41} - \f{\Delta \tilde{t}_{64}}{\Delta \tilde{t}_{53}} \Delta \tilde{t}_{64} \right) + f\left(\Psi\left(\f{p_4^\mathrm{f} + p_2^\mathrm{f}}{2} \right)\right), \label{eq:t_f_w_power}
 \end{align}
 where $p_i^\mathrm{f}$ is the FPP associated with timestamp $t_i$, and $\Psi(x) \triangleq 10^{(x - \alpha) / 10}$ is the lifting function suggested in \cite{Cano2022} with $\alpha$ as a normalization parameter. Moreover, $f(\cdot)$ is an unknown function to be learned from data, defined based on an experimentally-motivated assumption that the effects of power-correlated bias due to the individual tags can be aggregated into one function of the average received FPP that is common to all tags of similar design. The step (a) in the derivation involves the assumptions that $\rho_i\left(p_6^\mathrm{f}\right) = \rho_i\left(p_4^\mathrm{f}\right)$ and $\f{\Delta \tilde{t}_{64}}{\Delta \tilde{t}_{53}} \rho_j\left(p_2^\mathrm{f}\right) \approx \rho_j\left(p_2^\mathrm{f}\right)$. The former assumption is due to the fact that the motion of the robots is negligible in the time window $\Delta t_{64}$ and therefore the relative-pose between the two tags is similar, while the later assumption is due to $\rho_j\left(p_2^\mathrm{f}\right)$ being in the order of tenths of nanoseconds, and therefore $\left(1 - \f{\Delta \tilde{t}_{64}}{\Delta \tilde{t}_{53}}\right) \rho_j\left(p_2^\mathrm{f}\right) \approx 0$. 
 
 \begin{figure}
   \centering
    \includegraphics[trim={2cm 0cm 3cm 0cm},width=0.9\columnwidth]{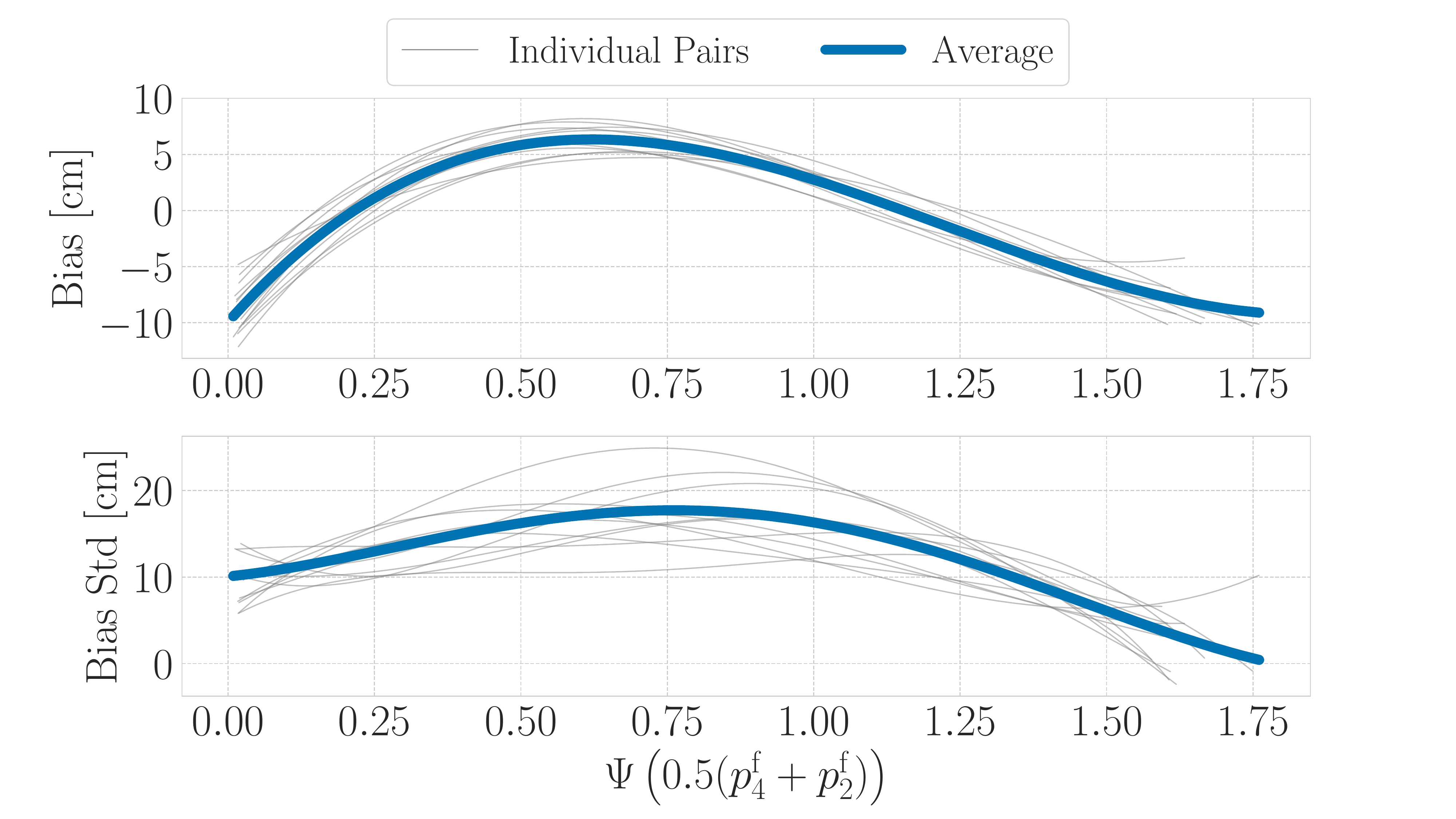}
    \caption{The fitted bias and standard deviation curves as a function of the lifted average FPP, using a 4-minute long training experiment.}
    \label{fig:bias_power_fit}
    \vspace{-12pt}
 \end{figure}

 Referring back to the experiment mentioned in Section~\ref{sec:antenna_delay_calibration}, the function $f$ is learned by simply fitting a spline to the post-antenna-delay-calibration range bias as a function of the lifted average FPP. The result is shown at the top of Figure~\ref{fig:bias_power_fit} for all individual pairs as well as for all the data, where the ToF bias is converted to range bias. As expected, when the antenna delay is corrected first, the bias-power curve is similar for all pairs as they all use the same antenna and PCB-board design. In this case, this additional calibration procedure can remove up to 10 cm of bias, but even though this calibration procedure is environment independent \cite{Cano2022}, this would vary for different tag designs and this process must be done separately for different antenna/board designs.
 
 \subsection{Variance Calibration}
 The variance of the measurements is also expected to vary as a function of the received-signal power, that is,
 \beq
    \mathbb{E}[(\tilde{t}_\mathrm{f}-t_\mathrm{f})^2] \triangleq \sigma\left(\Psi\left(\f{p_4^\mathrm{f} + p_2^\mathrm{f}}{2} \right)\right)^2. \label{eq:variance_power} \nonumber
 \eeq
 Intuitively, it is expected that the receiver should be able to detect and timestamp the direct-path signal more accurately when the FPP is high as this indicates a high signal-to-noise ratio (SNR). Additionally, multipath and obstacle-attenuated signals typically have lower FPP, and therefore not a lot of variance is expected at higher received FPP. At lower received FPP, the SNR is lower, and the received measurements might have been corrupted with equally-powerful multipath and body-attenuated signals. 
 
 In order to analyze this experimentally using the training data, a similar procedure to the power-bias calibration step is proposed. The standard-deviation samples are generated by computing the standard deviation of the range bias of the measurements in a window of FPP. A spline is fitted to the standard-deviation samples, and the resulting curves are shown at the bottom of Figure~\ref{fig:bias_power_fit}. As expected, the lowest standard deviation is at the highest FPP, where the standard deviation of the range bias is as low as 2.5 cm. Additionally, the highest standard deviation of approximately 17 cm is in the mid-FPP region. This is potentially due to reflections off the ground being primarly in this region.  
 
 Even though there is a clear trend, the standard deviation curves appear to somewhat vary between different pairs. This is partly due to the training experiment being relatively short, and at some FPP values there is not enough data points to accurately compute the standard deviation. Additionally, despite the curve seeming to slope downwards at lower FPP values, it is expected that at some point as the received FPP value decreases beyond the lower detection threshold the variance will increase drastically. However, the lower detection threshold is chosen to be higher than the point where random meaningless signals would be detected, and therefore the point where the variance increases significantly does not appear in the recorded experiment.

% \section{NLOS and Outlier Detection}

% convex hulls for NLOS detection using mocap

% show effect of NLOS on bias on training data

% NIS?

\section{Experimental Results on Testing Data} \label{sec:experimental_results}

The proposed calibration procedure from Sections \ref{sec:antenna_delay_calibration} and \ref{sec:power_correlated_calibration} is evaluated on 2 testing experiments with the same set-up as the training experiment presented in Section \ref{subsec:antenna_delay_training_exp}. While in the training experiment the quadcopters follow a more structured trajectory, the testing experiments involve the quadcopters flying around the 3-dimensional space randomly. Each testing experiment consists of 60 seconds of flight time and 10000 range measurements between the 12 pairs.

\subsection{Bias Correction} \label{subsec:bias_testing}

\begin{figure}[t!]
   \centering
   \includegraphics[trim={3cm 0cm 2cm 0cm},width=0.85\columnwidth]{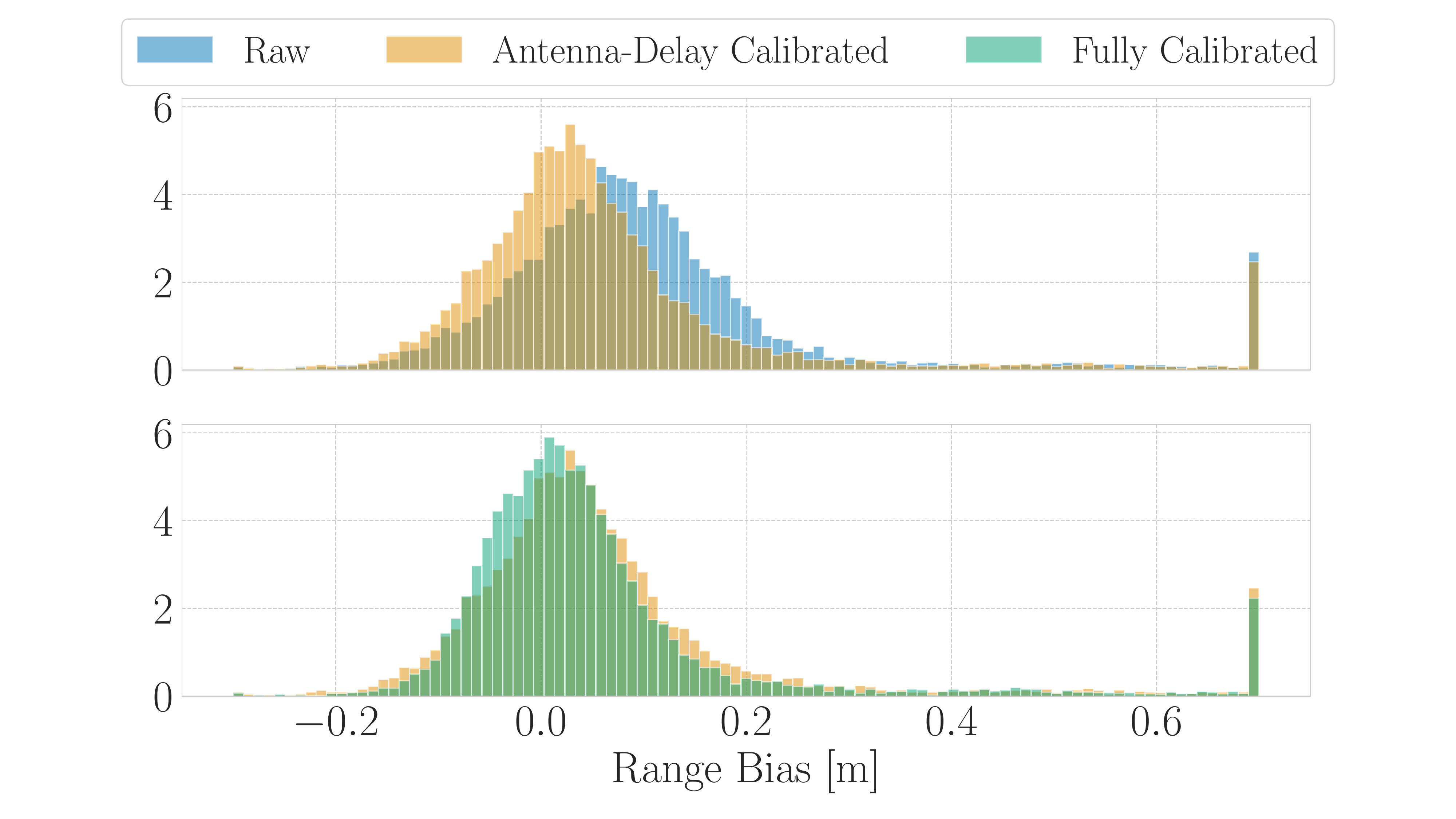}
   \caption{Distribution of the testing-data range bias pre- and post-calibration, with collection bins indicating the amount of outliers beyond the axes.}
   \label{fig:test_data_histograms}
   \vspace{-10pt}
\end{figure}

The distributions of the resulting biases in the testing data pre- and post-calibration are shown in Figure~\ref{fig:test_data_histograms}. The mean of the bias of the raw measurements is reduced by 36\% through antenna-delay calibration only and by a further 20\% by fully calibrating the measurements, bringing the mean from 11.11 cm to 5.91 cm. The standard deviation of the measurements is barely affected by antenna-delay calibration, but is reduced approximately 6\% through power-correlated calibration from 18.95 cm to 17.82 cm. Both the mean and standard deviation are affected by positive outliers potentially resulting from non-LOS and multipath propagation. 

In order to reject outliers, the underlying distribution must be known. Through the variance calibration procedure, the range measurements are assumed to be corrupted with zero-mean Gaussian noise with a standard deviation given by \eqref{eq:variance_power}. An outlier can be rejected if it does not satisfy the underlying distribution with a certain degree of confidence. For each individual measurement $k$ with ground-truth-computed bias $b_k$ and power-correlated standard deviation $\sigma_k$ from \eqref{eq:variance_power}, this can be done by performing the \emph{chi-squared test} \cite[Section 1.4.17]{barshalom2002}. Any measurement that does not satisfy the inequality $\f{b_k^2}{\sigma_k^2} \leq \gamma$ indicates that it is not from the underlying distribution to a certain degree of confidence. The threshold $\gamma$ depends on the chosen degree of confidence, typically 95\%. 

\begin{figure}[t!]
   \centering
   \includegraphics[trim={3cm 0cm 3cm 0cm},width=0.85\columnwidth]{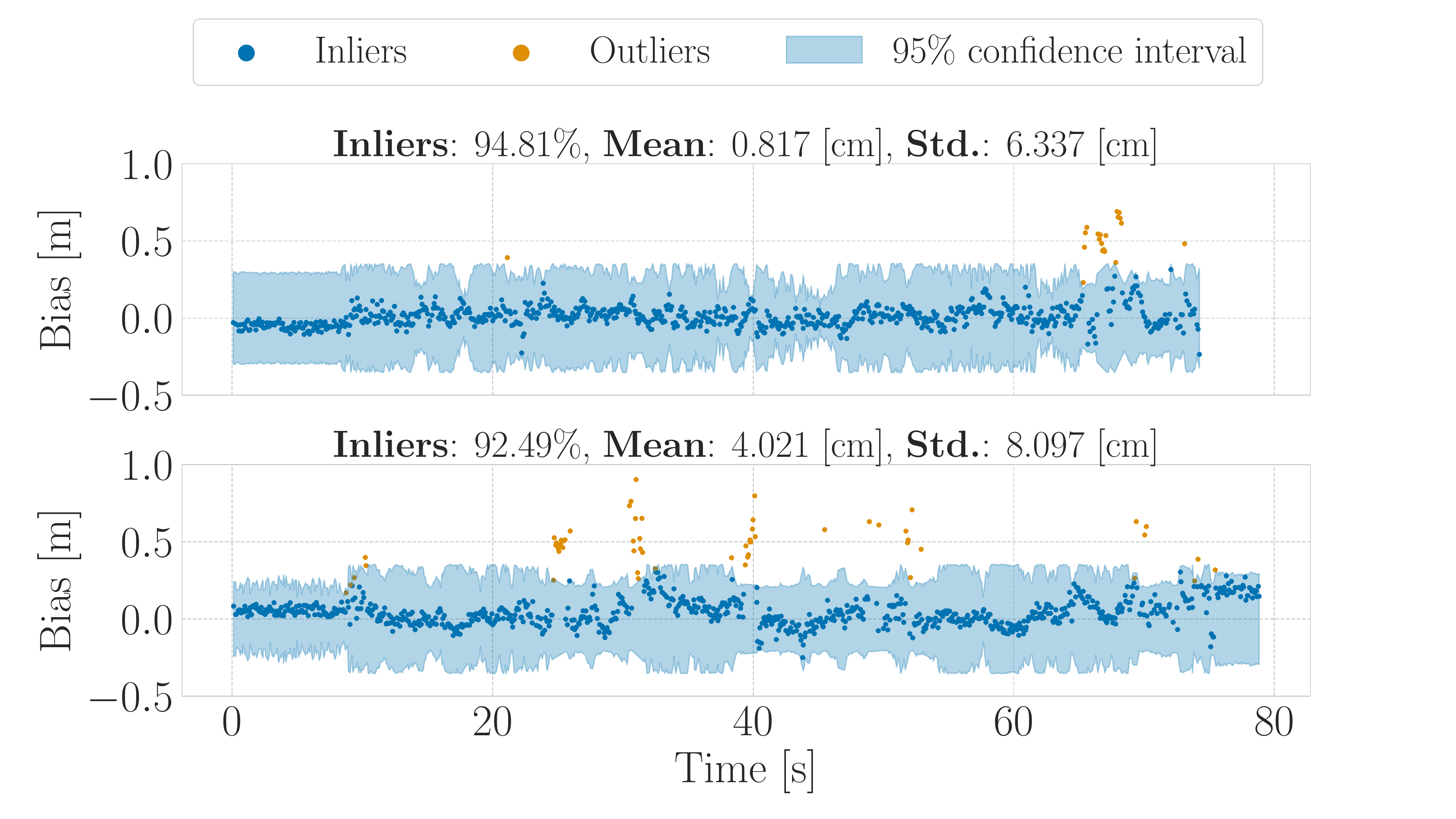}
   \caption{The range bias for 2 ranging pairs in one testing experiment after calibration, with some statistical information.}
   \label{fig:test_data_calibrated_measurements}
   \vspace{-11pt}
\end{figure}

The results of this standard outlier-rejection algorithm are shown in Figure \ref{fig:test_data_calibrated_measurements}. An indication of the calibrated variance being close to the actual underlying distribution would be that exactly 5\% of measurements are rejected; however, due to other factors such as non-LOS and multipath propagation, more than 5\% of measurements are rejected for some pairs. Even in this non-ideal scenario with non-LOS and multipath, the mean of the measurements after outlier rejection among all pairs reduces to a maximum of roughly 4 cm and a maximum standard deviation of approximately 8 cm.   

\subsection{Position Estimator} \label{subsec:pos_estimator_testing}

 In real-world applications, the ground truth distance between tags is usually not known but is rather estimated, and the outlier rejection method is usually done using the \emph{normalized-innovation-squared} (NIS) test \cite[Section 5.4.2]{barshalom2002}. The NIS test is similar to the chi-squared test mentioned in Section \ref{subsec:bias_testing}, but additionally accommodates for uncertainty in the state estimates. 
 
 To evaluate the variance calibration using the NIS test, the following simple localization problem is formulated using the testing data. Consider the problem of estimating $\mbf{r}_a^{1w}$, the position of Robot 1 relative to some arbitrary reference point $w$, resolved in some inertial frame $\mathcal{F}_{a}$. There are two tags on the robot, and $\mbf{r}_1^{i1}$ represents the position of Tag $i$ relative to the robot's reference point in the robot's own body frame $\mathcal{F}_{1}$, which can be measured manually. Additionally, assume that the orientation of the robot given by a direction cosine matrix $\mbf{C}_{1a} \in SO(3)$ is known, and that velocity measurements $\mbf{v}_a^{1w}$ are available from the motion-capture system. Assuming that poses and tag positions of the neighbouring robots $n \in \{ 2, 3 \}$ are known, an extended Kalman filter (EKF) is used to estimate $\mbf{r}_a^{1w}$, where the measurements are modelled as 
 \beq
    y = \norm{\mbf{r}_a^{1w} + \mbf{C}_{1a}^\trans \mbf{r}_1^{i1} - \mbf{r}_a^{nw} - \mbf{C}_{na}^\trans \mbf{r}_n^{jn}} + \nu \nonumber
 \eeq
 for the range measurement between Tag $i$ on Robot 1 and Tag $j$ on Robot $n$, and $\nu \sim \mathcal{N}\left( 0,R \right)$ is white Gaussian noise. The NIS test is used in the filter for outlier rejection. 
 
 \begin{figure}
   \centering
   \includegraphics[trim={0cm 0cm 0cm 0cm},width=0.85\columnwidth]{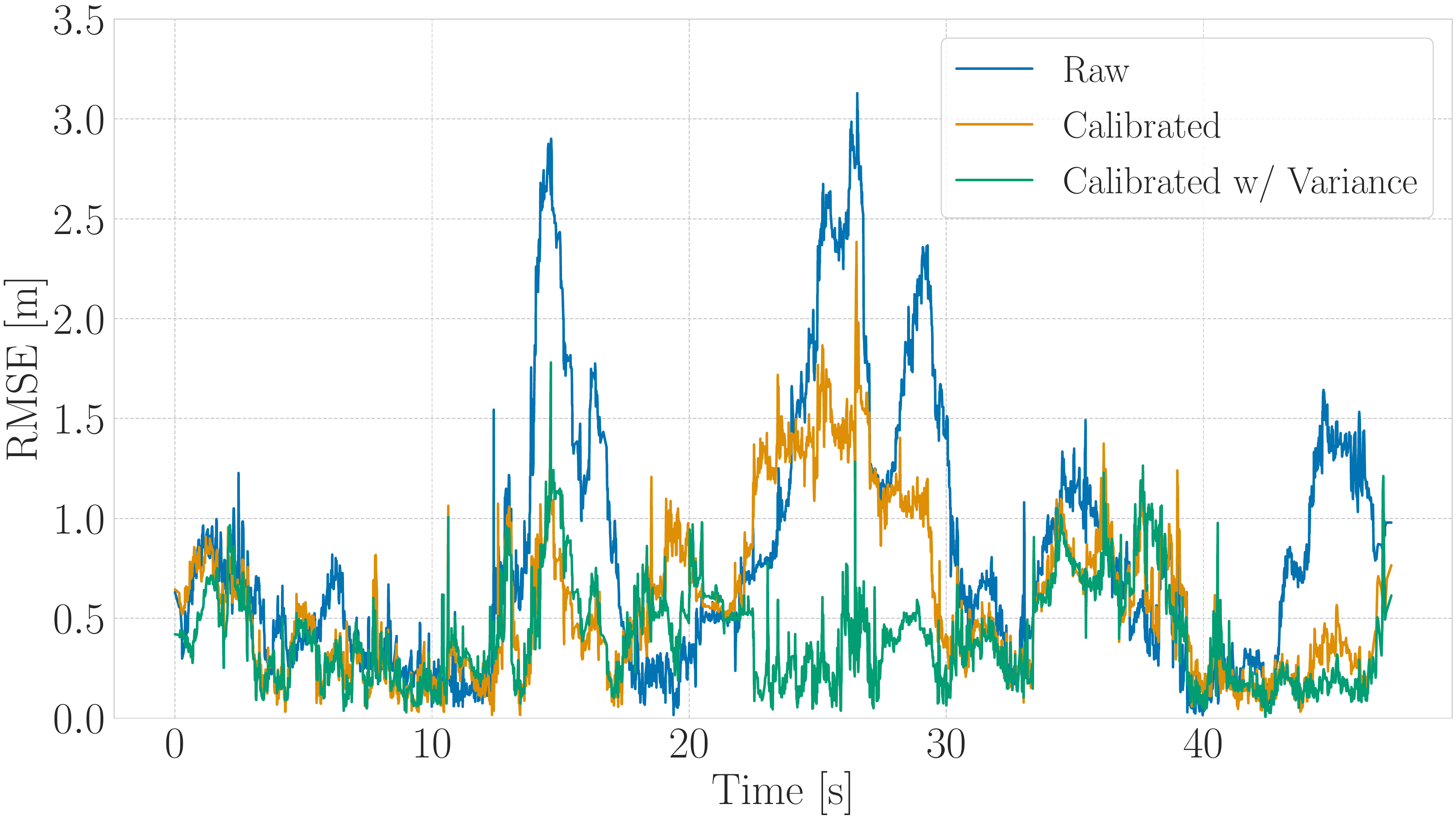}
   \caption{Comparison of the position-estimate RMSE using raw and calibrated data for one testing-data scenario.}
   \label{fig:pos_estimator_error_norm}
   \vspace{-5pt}
\end{figure}
 
%  \begin{table}
%     \centering
%     \caption{RMSE comparison for the raw and calibrated measurements in all 6 training-data scenarios. \textcolor{red}{Reformat.}}
%     \label{tab:testing_rmse}
%     \begin{tabular}{ p{1.5cm}|p{1.5cm}||c|c|p{1.5cm}} 
%      & \textbf{Robot No.} & \multicolumn{3}{c}{\textbf{RMSE [m]}} \\
%      \hline 
%      &  & \textbf{Raw} & \textbf{Calibrated} & \textbf{Calibrated w/ Variance} \\ 
%     \hline\hline
%     \multirow{3}{*}{\textbf{Experiment 1}} & 1 & 0.9245 & 0.5122 & \textbf{0.4243} \\ 
%       & 2 & 0.7577 & 0.5095 & \textbf{0.4437} \\ 
%       & 3 & 0.6731 & 0.3872 & \textbf{0.3467} \\ 
%     \hline 
%     \multirow{3}{*}{\textbf{Experiment 2}} & 1 & 0.8664 & \textbf{0.4879} & 0.4932 \\ 
%       & 2 & 0.8174 & 0.5854 & \textbf{0.3980} \\ 
%       & 3 & 0.7837 & 0.5167 & \textbf{0.4788} \\ 
%     \end{tabular}
%  \end{table}
 
\begin{table}[]
   \centering
   \caption{RMSE comparison for the raw and calibrated measurements in all 6 testing-data scenarios.}
   \label{tab:testing_rmse}
   \begin{tabular}{@{}ccccc@{}}
    & \textbf{} & \multicolumn{3}{c}{\textbf{RMSE {[}m{]}}} \vspace{3pt} \\
    & \multicolumn{1}{c}{\textbf{\begin{tabular}[c]{@{}c@{}}Robot\\ no.\end{tabular}}} & \textbf{Raw} & \textbf{Calibrated} & \textbf{\begin{tabular}[c]{@{}c@{}}Calibrated \\ w/ Variance\end{tabular}} \\ \midrule \midrule
   \multicolumn{1}{c}{\multirow{3}{*}{\textbf{Experiment 1}}} & \multicolumn{1}{c}{1} & 0.9245 & 0.5122 & \textbf{0.4243} \\
   \multicolumn{1}{c}{} & \multicolumn{1}{c}{2} & 0.7577 & 0.5095 & \textbf{0.4437} \\
   \multicolumn{1}{c}{} & \multicolumn{1}{c}{3} & 0.6731 & 0.3872 & \textbf{0.3467} \\ \midrule
   \multicolumn{1}{c}{\multirow{3}{*}{\textbf{Experiment 2}}} & \multicolumn{1}{c}{1} & 0.8664 & \textbf{0.4879} & 0.4932 \\
   \multicolumn{1}{c}{} & \multicolumn{1}{c}{2} & 0.8174 & 0.5854 & \textbf{0.3980} \\
   \multicolumn{1}{c}{} & \multicolumn{1}{c}{3} & 0.7837 & 0.5167 & \textbf{0.4788}
   \end{tabular}
   \vspace{-12pt}
\end{table}

 The performance of the filter is shown based on the \emph{root-mean-squared-error} (RMSE) metric in Figure~\ref{fig:pos_estimator_error_norm} for one experimental run and summarized in Table~\ref{tab:testing_rmse} for 6 different scenarios using 1) the raw measurements and fixed variance, 2) the calibrated measurements and fixed variance, and 3) the calibrated measurements and the calibrated variance-power curve. The 6 scenarios represent a variation of which of the three robots is the one with an unknown position to be estimated, and doing so in two different experimental runs. The choice of fixed measurement variance is decided experimentally based on what consistently yields the best performance. On average for the 6 different scenarios, the antenna-delay and bias-calibration procedures alone yield a 38\% improvement in localization accuracy, while additionally utilizing the power-correlated variance calibration results in an average of 46\% reduction in the RMSE, thus emphasizing the importance of calibrating UWB sensors and the added benefit of using the received FPP as an indication of the uncertainty of measurements.

\section{Conclusion}

In this paper, the problem of calibrating UWB bias is addressed. To eliminate the need for estimating the clock states, a DS-TWR-based ranging protocol is presented and shown to theoretically mitigate the effect of clock-skew-dependent bias. Furthermore, a robust and scalable antenna-delay calibration procedure is presented and trained on data from an aerial experiment. A model is then learnt on the experimental data to find the relation between the remaining bias and the uncertainty of the measurements as a function of the received-signal power. The delays and models learnt are then applied to two testing experiments to evaluate the calibration procedure. A localization problem is then formulated using an EKF, and it is shown that an average of 46\% improvement in localization accuracy can be achieved by using the corrected measurements and the modelled variance.   

%%%%%%%%%%%%%%%%%%%%%%%%%%%%%%%%%%%%%%%%%%%%%%%%%%%%%%%%%%%%%%%%%%%%%%%%%%%%%%%%
% \bibliographystyle{IEEEtran}
% \bibliography{IEEEabrv,IEEEexample}
{\AtNextBibliography{\small}
\printbibliography}
\end{document}